 \documentclass[pmlr]{jmlr}

 \RequirePackage{graphicx}
  \usepackage{booktabs}
 \usepackage{longtable}
 \makeatletter
 \def\set@curr@file#1{\def\@curr@file{#1}} 
 \makeatother
 \usepackage[load-configurations=version-1]{siunitx} 
 
 \usepackage{hyperref}
 \usepackage{url}
 \usepackage{wrapfig}
 \usepackage[utf8]{inputenc} 
 \usepackage{comment}
 \usepackage{caption}
 \usepackage{subcaption}
 \usepackage{pifont}

 \theorembodyfont{\upshape}
 \theoremheaderfont{\scshape}
 \theorempostheader{:}
 \theoremsep{\newline}

 \jmlrvolume{}
 \jmlryear{}
 \jmlrworkshop{}
 
 \title[Tree of Concepts]{Tree of Concepts: Interpretable Continual Learners in Non-Stationary Clinical Domains}
 
\author{\Name{Dongkyu Cho}
       \Email{dongkyu.cho@nyu.edu}\\
       \addr Computer Science Department\\
       \addr New York University\\
       New York, NY, USA
       \AND
       \Name{Xiyue Li}
       \Email{xiyue.li@nyulangone.org}\\
       \addr Department of Population Health\\
       NYU Grossman School of Medicine\\
       New York, NY, USA
       \AND
       \Name{Samrachana Adhikari}
       \Email{samrachana.adhikari@nyulangone.org}\\
       \addr Department of Population Health\\
       NYU Grossman School of Medicine\\
       New York, NY, USA
       \AND
       \Name{Rumi Chunara}
       \Email{rumi.chunara@nyu.edu}\\
       \addr School of Global Public Health\\
       \addr Computer Science Department\\
       New York University\\
       New York, NY, USA}

\begin{document}
 
 \maketitle
 
 \begin{abstract}
Continual learning aims to update models under distribution shift without forgetting, yet many high-stakes deployments, such as healthcare, also require interpretability. In practice, models that adapt well (e.g., deep networks) are often opaque, while models that are interpretable (e.g., decision trees) are brittle under shift, making it difficult to achieve both properties simultaneously. In response, we propose Tree of Concepts, an interpretable continual learning framework that uses a shallow decision tree to define a fixed, rule-based concept interface and trains a concept bottleneck model to predict these concepts from raw features. Continual updates act on the concept extractor and label head while keeping concept semantics stable over time, yielding explanations that do not drift across sequential updates. On multiple tabular healthcare benchmarks under continual learning protocols, our method achieves a stronger stability-plasticity trade-off than existing baselines, including replay-enhanced variants. Our results suggest that structured concept interfaces can support continual adaptation while preserving a consistent audit interface in non-stationary, high-stakes domains.
 
 \end{abstract}
 
 \section{Introduction}\label{sec:introduction}
 Machine learning systems deployed in practice operate under distributional change. Data distributions evolve across time, deployment sites, and user populations in ways that cannot be fully anticipated during training. Continual learning has emerged as a central paradigm to address this, aiming to adapt models to new data (plasticity) while preserving previously acquired knowledge (stability)~\citep{kirkpatrick2017overcoming}. Most successful continual-learning approaches rely on deep models that adapt effectively through fine-tuning or online updates. At the same time, many deployment settings impose a second requirement: interpretability~\citep{rymarczyk2023icicleinterpretableclassincremental}. In high-stakes domains such as healthcare, finance, and public policy, models must not only perform well but also expose reasoning that can be understood and audited by humans~\citep{song2025mindgapexaminingselfimprovement,cho2025correct}. In clinical use, this requirement is especially important because predictions can influence high-consequence decisions.

These requirements are often in tension. Interpretable models are typically brittle under distribution shift and difficult to update continually. Decision trees, for example, can be sensitive to changes in feature distributions and often require retraining when new data arrives. Deeper models, conversely, learn robust representations that support adaptation, but do so through latent variables that are difficult to interpret. This tension is especially apparent in clinical settings, where shifts are common and safety requirements are strict~\citep{goetz2024generalization}. Existing approaches usually optimize one side of this trade-off: continual adaptation without interpretability, or interpretability with weaker robustness under shift.

In this work, we argue that the trade-off is not fundamental. We introduce \textbf{Tree of Concepts}, a framework that reconciles continual learning and interpretability by decoupling representation learning from decision logic. The core idea is to treat decision-tree leaves as a fixed, rule-defined concept vocabulary~\citep{koh2020conceptbottleneckmodels,yu2025languageguidedconceptbottleneck}, and train a concept extractor to predict these concepts from raw features. Predictions are produced by a trainable concept-to-label head attached to this fixed concept interface. Because the concept vocabulary is fixed, explanation semantics remain stable across updates even as trainable components adapt to new slices. We evaluate on UCI Heart Disease~\citep{heart_disease_45}, CDC Diabetes Health Indicators~\citep{xie2019building}, and MIMIC-III~\citep{johnson2016mimic} under continual-learning protocols. Across datasets, Tree of Concepts improves the stability--plasticity trade-off over strong baselines, including replay-enhanced variants~\citep{rolnick2019experience}.

\paragraph{Generalizable Insights about Machine Learning in the Context of Healthcare}

The results of this work highlight a general lesson for clinical machine learning. Clinical datasets routinely exhibit distribution shift and cohort heterogeneity, and these effects often intensify under sequential model updates. In this regime, optimizing plasticity with high-capacity latent representations can conflict with interpretability and repeatable auditing. Our findings suggest that keeping the explanation interface fixed while adapting internal predictive mappings is a practical way to balance adaptation and transparency in non-stationary, safety-critical settings.
 
 \vspace{-3mm}
 \section{Related Works}\label{sec:related_works}
 \paragraph{Continual learning.}  
Continual learning studies how to update models on sequential tasks or drifting streams while limiting catastrophic forgetting~\citep{kirkpatrick2017overcoming}.
Representative families include regularization methods that protect parameters important for past tasks~\citep{kirkpatrick2017overcoming, aljundi2018memory, kang2022forget}, rehearsal methods that retain and replay exemplars~\citep{rebuffi2017icarlincrementalclassifierrepresentation}, and expansion-based approaches that add capacity to trade off stability and plasticity~\citep{der, foster}.
Rehearsal is a standard strong baseline, but performance depends on buffer size and replay design~\citep{rebuffi2017icarlincrementalclassifierrepresentation, zhou2024class}, and replay-based training can dominate compute relative to raw storage in some regimes~\citep{prabhu2023computationally, chavan2023towards, Harun_2023_CVPR, cho2026forgetforgettingcontinuallearning}.
Much of the literature targets deep architectures in vision and language, whereas tabular continual learning is comparatively less developed even though structured covariates are ubiquitous in operational systems. Recent studies examine continual learning for tabular data and show that mixed feature types, sparse interactions, and strong tree-based baseline can yield a different stability--plasticity profile than in perceptual domains~\citep{lourenço2025bridgingstreamingcontinuallearning, garcíasantaclara2024overcomingcatastrophicforgettingtabular}.
The gap matters for clinical risk modeling under temporal, demographic, and site shift.
We highlight medicine as an underdeveloped but high-impact area for continual learning, and hospital-cohort studies show that replay plus regularization can improve retention for risk models~\citep{amrollahi2022leveraging}; however, most clinical continual learning still emphasizes deep tabular predictors~\citep{lee2020clinical} or imaging~\citep{qazi2026continual}, with little focus on interpretable learners for non-stationary clinical tabular data.

\paragraph{Concept bottleneck models.}
Concept bottleneck models (CBMs) were introduced as a framework that predicts human-interpretable concepts before predicting the final label, enabling concept-level explanations and test-time interventions while retaining competitive task performance~\citep{koh2020conceptbottleneckmodels}. Since then, a central issue in the CBM literature has been the concept labeling bottleneck: standard CBMs typically assume a predefined concept vocabulary together with dense concept annotations, which are expensive to obtain and difficult to scale. Several subsequent works address this limitation from different angles. Post-hoc CBMs show that a pretrained black-box model can be converted into a concept-based predictor after training, greatly reducing the performance penalty usually associated with concept supervision~\citep{yuksekgonul2023posthocconceptbottleneckmodels}. Label-free CBMs go further by removing the need for labeled concept data and automatically aligning model representations with interpretable concepts~\citep{oikarinen2023labelfreeconceptbottleneckmodels}, while Discover-then-Name relaxes the need to even specify the concept set in advance by automatically discovering and naming concepts learned by the model~\citep{rao2024discoverthennametaskagnosticconceptbottlenecks}. More recent work has continued to extend CBMs toward practical settings, including clinical knowledge-guided variants in medicine and continual-learning variants that aim to preserve semantic consistency of concepts across tasks~\citep{pang2024integratingclinicalknowledgeconcept, rymarczyk2023icicleinterpretableclassincremental, yu2025languageguidedconceptbottleneck}. However, these approaches are largely developed in visual or class-incremental settings, and they often rely on manually curated or externally aligned concepts. In contrast, our setting is non-stationary clinical tabular learning, where we derive the bottleneck from shallow decision-tree rules over structured features, yielding a fixed and human-auditable concept vocabulary that remains semantically stable across continual updates.
 
 \vspace{-2mm}
 \section{Problem Formulation}\label{sec:theory}
 In this section, we define the core problem addressed in this paper: interpretable continual learning under distribution shift. Specifically, we study supervised learning over a non-stationary stream of data slices
$D_1,\dots,D_T$, where each slice
$D_t=\{(x_i^{(t)},y_i^{(t)})\}_{i=1}^{n_t}$ is drawn from a shifted distribution.
At step $t$, the learner updates a model $h_t$ using $D_t$ (and optional replay data)
without retraining from scratch.

Continual learning favors models that can adapt to new slices while retaining past knowledge.
Interpretability, in contrast, oftn favors shallow models with fixed and human-readable logic.
In practice, models that adapt well are often opaque, while models that are easy to audit are brittle under shift.
Our goal is to satisfy both requirements at once.

\paragraph{Objective.}
For each step $t$, we want high \emph{plasticity} on current data and high \emph{stability} on previous data:
\begin{align}
\text{Plasticity}_t &= \mathcal{P}_t := \mathcal{M}(h_t, D_t),\\
\text{Stability}_t  &= \mathcal{S}_t := \frac{1}{t-1}\sum_{j=1}^{t-1}\mathcal{M}(h_t, D_j),
\end{align}
where $\mathcal{M}$ is a task metric (e.g., AUROC, accuracy).
We seek models with simultaneously high $\mathcal{P}_t$ and $\mathcal{S}_t$ over the full sequence.

Beyond performance, we argue that predictions must be explained through a stable, rule-auditable interface: the semantics of explanation units should remain consistent across updates.
Formally, if $e_t(x)$ denotes the explanation produced at step $t$, we require semantic consistency
over time (no drifting meaning of explanation units), while still balancing the stability-plasticity tradeoff.
 
 \vspace{-3mm}
 \section{Proposed Method: Tree of Concepts}\label{sec:method}
 \subsection{Notation}
We consider supervised learning under a nonstationary data stream segmented into slices
$D_1,\dots,D_T$, where slice $t$ provides
$D_t=\{(x_i^{(t)},y_i^{(t)})\}_{i=1}^{n_t}$ with features $x\in\mathbb{R}^d$ and labels
$y\in\{1,\dots,K\}$. Slices may index time windows, deployment sites, or evolving patient cohorts.
We write $\mathcal{R}_t(\theta)=\mathbb{E}_{(x,y)\sim D_t}[\ell(\hat y_\theta(x),y)]$ for the risk on slice $t$.
Our goal is to learn a sequence of models that adapts to new slices while avoiding performance regressions on previously
seen slices, and exposes a stable, rule-auditable rationale for each prediction.

\subsection{Main idea}

We seek continual learning under distribution shift while maintaining an interpretable decision interface suitable for high risk deployment. For this, we define a fixed set of interpretable concepts using a shallow decision tree, and train a concept bottleneck model that
predicts these concepts from raw features and then predicts labels from the resulting concept representation. Continual updates act on the trainable concept extractor and concept to label predictor, while the tree-defined concepts remain
fixed, yielding explanations whose semantics do not drift across slices.

\subsection{Decision Trees for robust concept extraction}
We train a shallow decision tree $\mathcal{T}$ on an initial broad slice (or an early pooled window) and treat its
leaf nodes as a discrete concept vocabulary. Each leaf corresponds to a path rule, a conjunction of feature threshold tests,
which yields a concise, human-readable description of the region of feature space represented by that leaf.

The decision tree serves as an interpretable scaffold that provides reliable and auditable concepts without requiring
manual concept annotation. Importantly, we freeze $\mathcal{T}$ after initialization so that each concept retains a fixed
meaning over time, which is critical when the model is updated continually.

\paragraph{Concept targets.}
Let $\{1,\dots,L\}$ denote the leaf set of $\mathcal{T}$. For any input $x$, routing through the tree yields a leaf index
\begin{equation}
  z(x)=\mathrm{leaf}_{\mathcal{T}}(x)\in\{1,\dots,L\}.
\end{equation}
We use $z(x)$ as a pseudo label for concept learning in every slice. Since $\mathcal{T}$ is fixed, concept supervision
is available for both current and replayed data without additional labeling.

\subsection{Concept Bottleneck Models for continual adaptation}

We instantiate a two-stage predictor consisting of a concept extractor and a label head. The concept extractor learns a
robust mapping from raw features to the tree-defined concept space, and the head learns the downstream mapping from concepts
to labels.

Decision trees provide transparent concepts but impose hard partitions that can be brittle under covariate shift~\citep{moshkovitz2021connecting}.
The concept bottleneck model~\citep{koh2020conceptbottleneckmodels} complements the tree by learning a soft, trainable concept predictor that can generalize across
shifted feature distributions while remaining anchored to the fixed concept vocabulary. The label head isolates the
concept to outcome mapping, which can be updated as clinical correlations evolve, without changing the concept semantics.

\paragraph{Model.}
We define a concept predictor (LeafNet) $f_\phi:\mathbb{R}^d\rightarrow \Delta^{L}$ that outputs a distribution over
leaf concepts
\begin{equation}
  \hat c(x)=f_\phi(x), \qquad \hat c(x)\in\Delta^{L},
\end{equation}
and a label head $g_\psi:\Delta^{L}\rightarrow \Delta^{K}$ producing
\begin{equation}
  \hat y(x)=g_\psi(\hat c(x)).
\end{equation}
Using $\hat c(x)$ as a distribution supports robustness and uncertainty at the concept level; if a hard concept assignment is
preferred, we use $\hat z(x)=\arg\max_{\ell}\hat c_\ell(x)$ and pass a one-hot vector to the head.

\paragraph{Training objective.}
We train LeafNet to match the frozen tree concept assignment and train the head to predict the task label.
For a sample $(x,y)$ we optimize
\begin{align}
  \mathcal{L}_{\text{concept}}(x)
  &= \mathrm{CE}\!\left(f_\phi(x),\, z(x)\right),\\
  \mathcal{L}_{\text{label}}(x,y)
  &= \mathrm{CE}\!\left(g_\psi(f_\phi(x)),\, y\right),\\
  \mathcal{L}(x,y)
  &= \mathcal{L}_{\text{concept}}(x) + \lambda\,\mathcal{L}_{\text{label}}(x,y),\label{eq:toc_total_loss}
\end{align}
where $\lambda$ controls the tradeoff between concept fidelity and end task performance.
The concept loss anchors the representation to the interpretable tree vocabulary, while the label loss drives task accuracy.

\subsection{Continual learning updates}
We freeze the tree $\mathcal{T}$ after the initial phase and update the trainable parameters $(\phi,\psi)$ sequentially
over slices. We maintain a bounded replay buffer $M$ containing samples from past slices. At slice $t$, each minibatch is
constructed by mixing current data and replayed data, for example by sampling
$B_t\subset D_t$ and $B_M\subset M$ and setting $B=B_t\cup B_M$.
We minimize $\sum_{(x,y)\in B}\mathcal{L}(x,y)$, computing $z(x)$ on the fly by routing through the frozen tree.
After training on $D_t$, we update $M$ by inserting a subset of $D_t$ (optionally balanced by label and/or concept index)
and evicting older items when capacity is exceeded. Replay stabilizes both concept prediction and concept-to-label mapping,
reducing forgetting while allowing adaptation to new regimes.

\paragraph{Summary}
Decision tree leaves provide a fixed set of rule-defined concepts, yielding a stable explanation vocabulary and automatic
concept supervision. The concept bottleneck model supplies continual learning capacity by learning a robust concept extractor
and an updatable concept to label head, both constrained by the fixed concept interface. This combination preserves an
auditable decision process while enabling continual updates under distribution shift.

 \vspace{-3mm}
 \section{Experiments}\label{sec:experiment}
 \subsection{Experimental setting}\label{sec:experimental-setting}
  
We evaluate Tree of Concepts under a \emph{continual slice} protocol: the model is trained sequentially on ordered subsets of the data that induce distribution shift, then tested on all subsets seen so far.
We report \emph{plasticity} as the task metric on the current slice and \emph{stability} as the mean of the same metric over all previously seen subsets.
Unless noted, all methods use the same split, features, and (when enabled) the same replay buffer size and mixing ratio between current and replayed mini-batches.

\paragraph{Open-access tabular datasets.}
We use two standard open-access datasets for reproducibility.
UCI Heart Disease~\citep{heart_disease_45} is binary classification (presence of heart disease) with the usual 14 attributes (13 inputs + label).
CDC Diabetes Health Indicators~\citep{xie2019building} is 3-class classification (healthy / pre-diabetes / diabetes) with 21 attributes (20 inputs + label).
For Heart Disease we partition by age into three slices $T_1{:}\ 18\text{--}34$, $T_2{:}\ 35\text{--}64$, $T_3{:}\ 65+$.
For CDC Diabetes we partition by age into four slices $18\text{--}34$, $35\text{--}49$, $50\text{--}64$, $65+$.

\paragraph{MIMIC-III cohort and task.}
We further evaluate on MIMIC-III (v1.4)~\citep{johnson2016mimic}, a de-identified ICU database with controlled access (credentialing and data-use approval required).
We formulate one prediction row per ICU stay with a binary in-hospital mortality label derived from admission/discharge information (hospital-expire indicator and death timestamp for the indexed admission).
The input is a \emph{fixed tabular schema} shared across all slices: static demographics (e.g., age, sex), admission and service context, comorbidity summaries encoded from ICD-9 (e.g., Elixhauser components or grouped counts), and early-trajectory summaries of labs and vitals from the first 24 hours after ICU admission (mean/min/max where applicable, plus \emph{missingness indicators} so shifts in measurement frequency do not change feature dimension).
We exclude future information: no variables measured predominantly after the prediction horizon are used as inputs.

\paragraph{MIMIC-III continual protocols.}
We study two complementary slice constructions that reflect realistic non-stationarity in critical care data.
(A) Time-window shift: slices are disjoint admission-year windows within MIMIC-III, trained in chronological order so later slices reflect evolving practice, coding, and cohort mix (exact year boundaries are listed in the appendix, Detailed Experimental Settings).
(B) Demographic shift: slices are disjoint patient cohorts by age bucket (e.g., 18--39, 40--64, 65--79, 80+), ordered from younger to older, holding the same label and feature definition.
In both protocols, we report AUROC for mortality; preprocessing (imputation rules, clipping, and aggregation windows) is frozen across slices.
Cohort size, slice-wise class prevalence, and train/validation/test counts are summarized in Appendix Table~\ref{tab:cohort_transparency}.

\paragraph{Baselines and metrics.}
Baselines are Decision Tree, XGBoost, and FT-Transformer, trained with the same slice schedule.
For \emph{replay} variants, we store a bounded buffer of past-slice examples and mix them into each update (matching capacity across methods for fair comparison).
Tree of Concepts uses replay for continual updates to the concept extractor and label head while keeping tree-defined concepts fixed (see Section~\ref{sec:method}).
We report: (i) full-data upper bound from training on all slices pooled (non-continual reference); (ii) Avg.\ Past-Task (stability); (iii) Avg.\ Current-Task (plasticity).
Open-access datasets use AUROC (Heart) and Macro-F1 (CDC); MIMIC uses AUROC.
Open-access experiments are repeated with five random seeds; MIMIC-III experiments are repeated with three random seeds.
We report mean $\pm$ standard error of the mean (SE) across runs.

\paragraph{Interpreting pooled vs.\ continual scores.}
The full-data upper bound uses a different training regime (joint access to all slices) than the continual metrics; absolute values are therefore not directly comparable across these columns.
We focus on relative ordering under a fixed protocol and on the stability-plasticity tradeoff (past vs.\ current task), where replay typically raises past-task performance with modest, method-dependent changes in current-task performance.

\subsection{Results}
We report results in two stages to match the validation trajectory of the paper: (i) open-access tabular datasets for controlled, reproducible comparison, then (ii) restricted-access MIMIC-III for higher-fidelity clinical evaluation under two distinct continual-shift protocols.

\paragraph{Open-access dataset results.}
Tables~\ref{tab:heart_results} and~\ref{tab:cdc_results} show that replay usually improves retention for baseline models, while Decision Tree remains the least stable option under sequential shift.
Tree of Concepts provides the strongest overall stability-plasticity balance on both datasets.

\begin{table}[t]
\centering
\small
\caption{UCI Heart Disease (binary classification, AUROC). We report the mean and standard error over five runs.}
\label{tab:heart_results}
\vspace{-1.5mm}
\footnotesize
\setlength{\tabcolsep}{3pt}
\begin{tabular}{lcccc}
\toprule
Method & Replay & Full-data UB & Avg.\ Past-Task & Avg.\ Current-Task \\
\midrule
\multicolumn{5}{l}{\textit{No Replay}} \\
Decision Tree & \ding{55} & $0.83 \pm 0.02$ & $0.56 \pm 0.04$ & $0.72 \pm 0.03$ \\
XGBoost & \ding{55} & $0.89 \pm 0.01$ & $0.75 \pm 0.03$ & $0.84 \pm 0.02$ \\
FT-Transformer & \ding{55} & $0.88 \pm 0.02$ & $0.73 \pm 0.03$ & $0.83 \pm 0.02$ \\
\midrule
\multicolumn{5}{l}{\textit{Replay}} \\
Decision Tree & \ding{51} & $0.83 \pm 0.02$ & $0.63 \pm 0.04$ & $0.73 \pm 0.03$ \\
XGBoost & \ding{51} & $0.89 \pm 0.01$ & $0.78 \pm 0.02$ & $0.85 \pm 0.02$ \\
FT-Transformer & \ding{51} & $0.88 \pm 0.01$ & $0.76 \pm 0.02$ & $0.84 \pm 0.03$ \\
\midrule
\multicolumn{5}{l}{\textit{Ours}} \\
Tree of Concepts & \ding{51} & $0.89 \pm 0.01$ & $0.80 \pm 0.02$ & $0.85 \pm 0.02$ \\
\bottomrule
\end{tabular}
\normalsize
\vspace{-1.5mm}
\end{table}

\begin{table}[t]
\centering
\small
\caption{CDC Diabetes Health Indicators (3-class classification, Macro-F1). We report the mean and standard error over five runs.}
\label{tab:cdc_results}
\vspace{-1.5mm}
\footnotesize
\setlength{\tabcolsep}{3pt}
\begin{tabular}{lcccc}
\toprule
Method & Replay & Full-data UB & Avg.\ Past-Task & Avg.\ Current-Task \\
\midrule
\multicolumn{5}{l}{\textit{No Replay}} \\
Decision Tree & \ding{55} & $0.58 \pm 0.02$ & $0.28 \pm 0.05$ & $0.47 \pm 0.04$ \\
XGBoost & \ding{55} & $0.65 \pm 0.02$ & $0.51 \pm 0.03$ & $0.60 \pm 0.02$ \\
FT-Transformer & \ding{55} & $0.64 \pm 0.01$ & $0.48 \pm 0.04$ & $0.57 \pm 0.03$ \\
\midrule
\multicolumn{5}{l}{\textit{Replay}} \\
Decision Tree & \ding{51} & $0.58 \pm 0.02$ & $0.35 \pm 0.05$ & $0.48 \pm 0.03$ \\
XGBoost & \ding{51} & $0.65 \pm 0.01$ & $0.55 \pm 0.03$ & $0.61 \pm 0.02$ \\
FT-Transformer & \ding{51} & $0.64 \pm 0.02$ & $0.52 \pm 0.03$ & $0.59 \pm 0.03$ \\
\midrule
\multicolumn{5}{l}{\textit{Ours}} \\
Tree of Concepts & \ding{51} & $0.66 \pm 0.01$ & $0.59 \pm 0.02$ & $0.63 \pm 0.02$ \\
\bottomrule
\end{tabular}
\normalsize
\vspace{-1.5mm}
\end{table}

\begin{figure*}[t]
\centering
\subfigure[UCI Heart Disease.]{
  \includegraphics[width=0.47\textwidth]{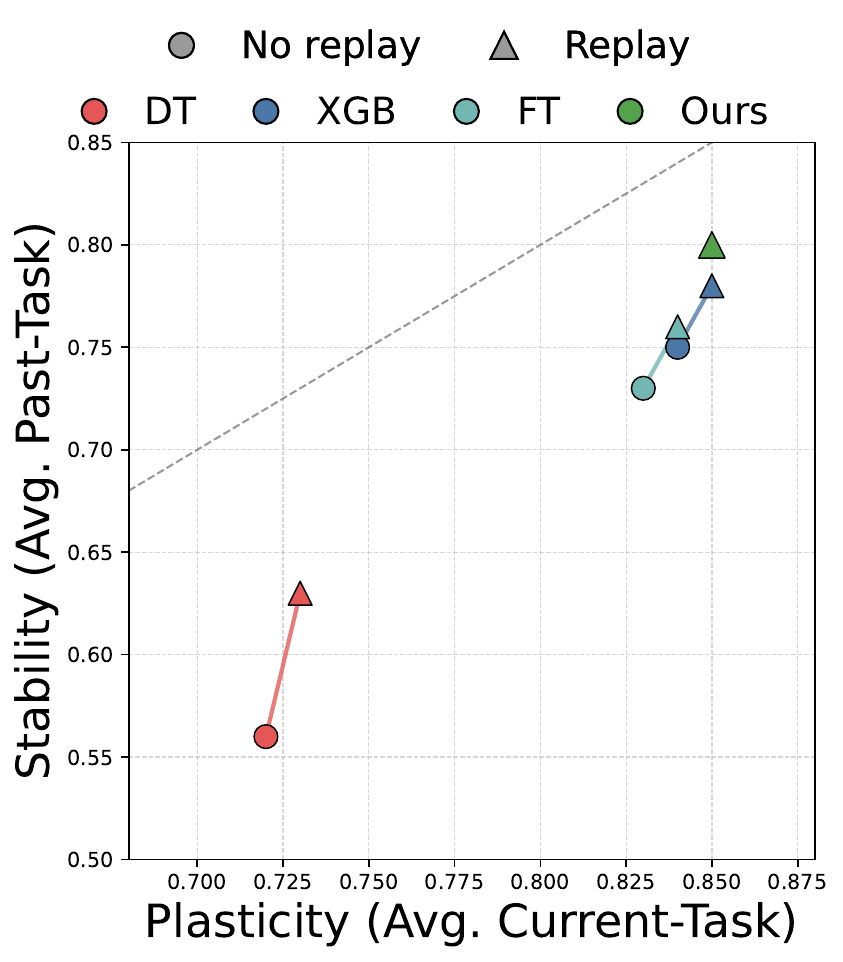}%
  \label{fig:heart_tradeoff}
}\hfill
\subfigure[CDC Diabetes Health Indicators.]{
  \includegraphics[width=0.47\textwidth]{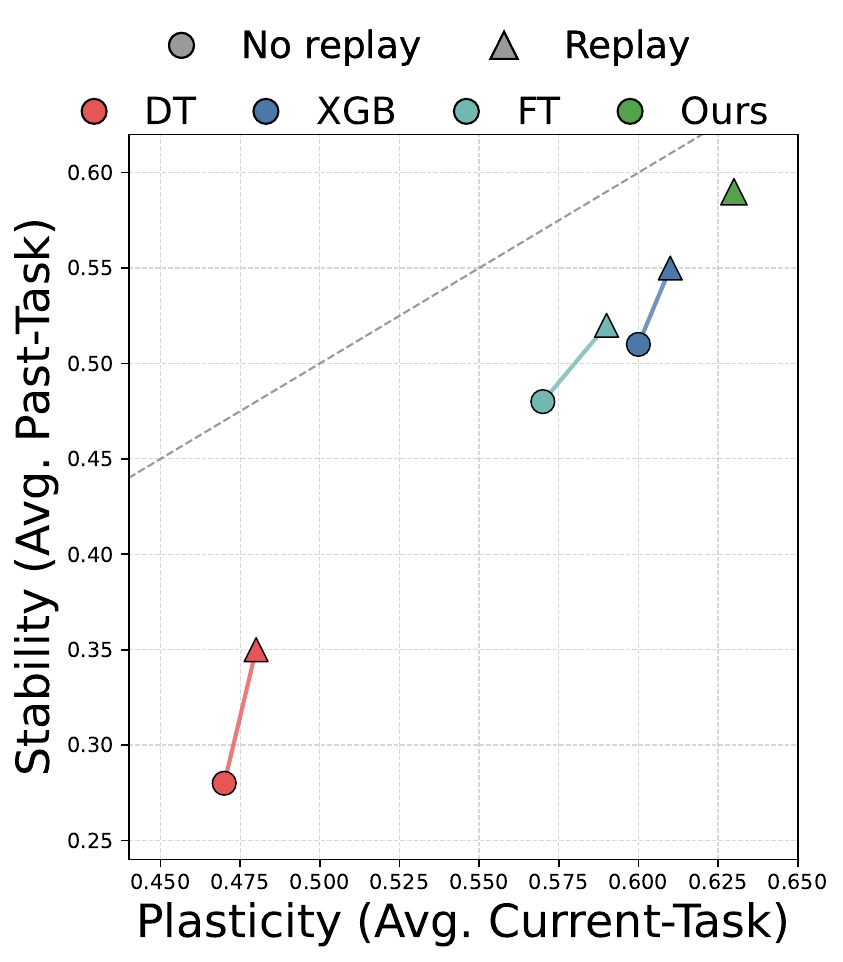}%
  \label{fig:cdc_tradeoff}
}
\caption{Stability--plasticity trade-off on open-access tabular datasets (mean across five runs; tables give mean $\pm$ SE). The x-axis is plasticity (Avg.\ Current-Task), and the y-axis is stability (Avg.\ Past-Task). Segments link no-replay and replay for each baseline.}
\label{fig:public_tradeoff_subplots}
\end{figure*}

\paragraph{MIMIC-III results.}
We then evaluate on MIMIC-III mortality prediction with two complementary shift definitions: protocol (A) time-window shift and protocol (B) demographic shift.
Protocol (A) probes temporal drift and practice change, while protocol (B) probes cohort shift across patient groups.
Reporting both helps distinguish whether gains are robust to different sources of non-stationarity rather than tied to a single slicing scheme.

\begin{table}[t]
\centering
\small
\caption{MIMIC-III in-hospital mortality (AUROC), continual protocol (A) time-window shift by admission year. We report the mean and standard error over three runs.}
\label{tab:mimic_time}
\vspace{-1.5mm}
\footnotesize
\setlength{\tabcolsep}{3pt}
\begin{tabular}{lcccc}
\toprule
Method & Replay & Full-data UB & Avg.\ Past-Task & Avg.\ Current-Task \\
\midrule
\multicolumn{5}{l}{\textit{No Replay}} \\
Decision Tree & \ding{55} & $0.78 \pm 0.02$ & $0.54 \pm 0.04$ & $0.70 \pm 0.03$ \\
XGBoost & \ding{55} & $0.85 \pm 0.01$ & $0.72 \pm 0.03$ & $0.79 \pm 0.02$ \\
FT-Transformer & \ding{55} & $0.84 \pm 0.02$ & $0.70 \pm 0.03$ & $0.78 \pm 0.03$ \\
\midrule
\multicolumn{5}{l}{\textit{Replay}} \\
Decision Tree & \ding{51} & $0.78 \pm 0.01$ & $0.61 \pm 0.04$ & $0.71 \pm 0.03$ \\
XGBoost & \ding{51} & $0.85 \pm 0.01$ & $0.76 \pm 0.02$ & $0.81 \pm 0.02$ \\
FT-Transformer & \ding{51} & $0.84 \pm 0.01$ & $0.73 \pm 0.03$ & $0.80 \pm 0.02$ \\
\midrule
\multicolumn{5}{l}{\textit{Ours}} \\
Tree of Concepts & \ding{51} & $0.86 \pm 0.01$ & $0.78 \pm 0.02$ & $0.82 \pm 0.02$ \\
\bottomrule
\end{tabular}
\normalsize
\vspace{-1.5mm}
\end{table}

\begin{table}[t]
\centering
\small
\caption{MIMIC-III in-hospital mortality (AUROC), continual protocol (B) demographic shift by age bucket. We report the mean and standard error over three runs.}
\label{tab:mimic_demo}
\vspace{-1.5mm}
\footnotesize
\setlength{\tabcolsep}{3pt}
\begin{tabular}{lcccc}
\toprule
Method & Replay & Full-data UB & Avg.\ Past-Task & Avg.\ Current-Task \\
\midrule
\multicolumn{5}{l}{\textit{No Replay}} \\
Decision Tree & \ding{55} & $0.77 \pm 0.01$ & $0.52 \pm 0.04$ & $0.68 \pm 0.03$ \\
XGBoost & \ding{55} & $0.84 \pm 0.02$ & $0.70 \pm 0.03$ & $0.77 \pm 0.02$ \\
FT-Transformer & \ding{55} & $0.83 \pm 0.01$ & $0.68 \pm 0.03$ & $0.76 \pm 0.02$ \\
\midrule
\multicolumn{5}{l}{\textit{Replay}} \\
Decision Tree & \ding{51} & $0.77 \pm 0.02$ & $0.58 \pm 0.04$ & $0.69 \pm 0.03$ \\
XGBoost & \ding{51} & $0.84 \pm 0.01$ & $0.74 \pm 0.03$ & $0.78 \pm 0.02$ \\
FT-Transformer & \ding{51} & $0.83 \pm 0.02$ & $0.71 \pm 0.03$ & $0.77 \pm 0.03$ \\
\midrule
\multicolumn{5}{l}{\textit{Ours}} \\
Tree of Concepts & \ding{51} & $0.85 \pm 0.01$ & $0.76 \pm 0.02$ & $0.80 \pm 0.02$ \\
\bottomrule
\end{tabular}
\normalsize
\vspace{-1.5mm}
\end{table}

\begin{figure*}[t]
\centering
\subfigure[Protocol (A): time-window shift.]{
  \includegraphics[width=0.47\textwidth]{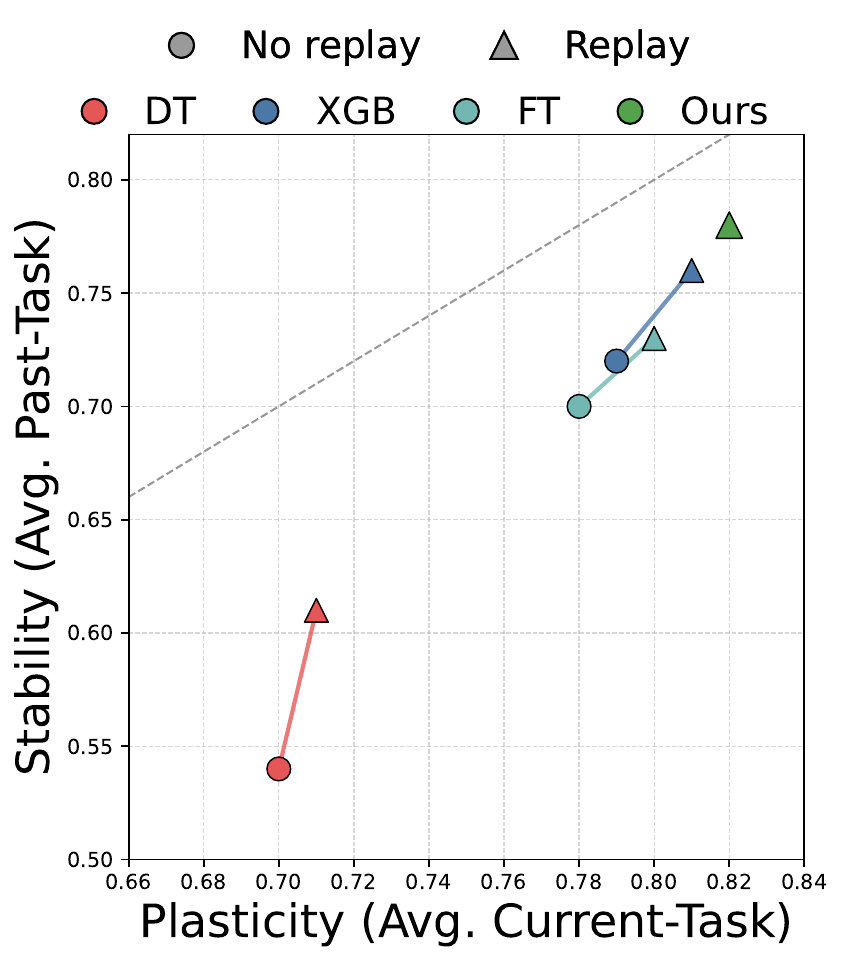}%
  \label{fig:mimic_time_tradeoff}
}\hfill
\subfigure[Protocol (B): demographic shift.]{
  \includegraphics[width=0.47\textwidth]{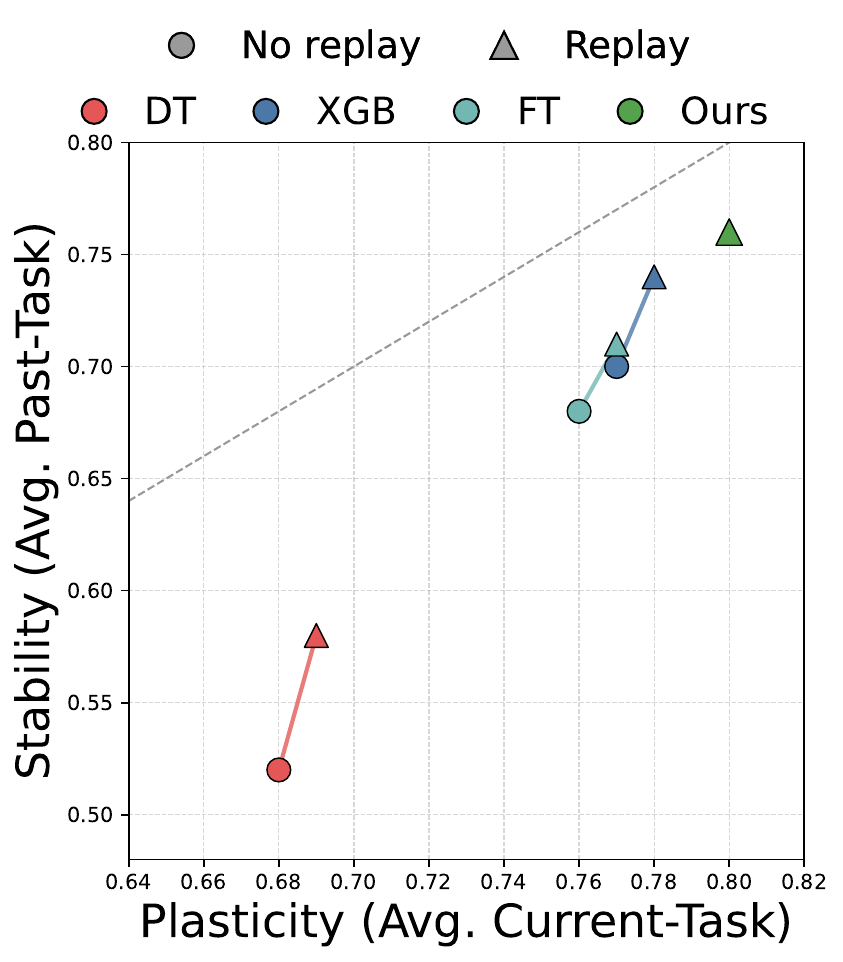}%
  \label{fig:mimic_demo_tradeoff}
}
\caption{Stability--plasticity trade-offs on MIMIC-III mortality prediction under protocols (A) and (B); points are means across three runs (see tables for $\pm$ SE).}
\label{fig:mimic_tradeoff_subplots}
\end{figure*}

\paragraph{Qualitative concept consistency and trustworthiness.}
The tree is frozen after initialization, so each concept ID always corresponds to the same rule.
At each slice, we compare (1) the concept ID assigned by the frozen tree, $z(x)$, and (2) the top concept ID predicted by the concept model, $\hat z(x)=\arg\max_{\ell}\hat c_{\ell}(x)$.
We report node agreement (higher is better), rule-fidelity gap (performance difference when using predicted concepts versus tree-assigned concepts; lower is better), and high-confidence contradiction rate (disagreement when confidence $>0.8$; lower is better).

\begin{table}[t]
\centering
\small
\caption{Concept consistency and trustworthiness of tree-node concepts extracted by the concept bottleneck model (mean $\pm$ SE; five runs for UCI/CDC and three runs for MIMIC-III).}
\label{tab:concept_consistency}
\vspace{-1.5mm}
\footnotesize
\setlength{\tabcolsep}{3pt}
\begin{tabular}{lccc}
\toprule
Setting & Node agreement $\uparrow$ & Rule-fidelity gap $\downarrow$ & High-conf.\ contradiction $\downarrow$ \\
\midrule
UCI Heart Disease & $0.91 \pm 0.01$ & $0.012 \pm 0.004$ & $0.039 \pm 0.009$ \\
CDC Diabetes & $0.86 \pm 0.02$ & $0.020 \pm 0.006$ & $0.056 \pm 0.011$ \\
MIMIC-III (A) Time-window & $0.83 \pm 0.02$ & $0.027 \pm 0.008$ & $0.070 \pm 0.013$ \\
MIMIC-III (B) Demographic & $0.81 \pm 0.02$ & $0.033 \pm 0.009$ & $0.078 \pm 0.014$ \\
\bottomrule
\end{tabular}
\normalsize
\vspace{-1.5mm}
\end{table}

Table~\ref{tab:concept_consistency} shows high agreement on UCI/CDC and lower agreement on MIMIC under stronger shift.
The fidelity gap remains small across all settings, indicating that predicted concepts stay close to the fixed tree interface even as predictive layers adapt.
Given space constraints, we report aggregate indicators here rather than per-rule case studies.

\paragraph{Observation.}
Replay usually improves Avg.\ Past-Task for tabular baselines across open-access datasets and MIMIC-III, while largely preserving or slightly improving current-slice performance, consistent with reduced forgetting.
These gains are obtained with a bounded replay memory and random exemplar sampling, indicating that a relatively small subset of past samples can already mitigate forgetting in our setting.
In every baseline block, the Decision Tree attains the lowest Avg.\ Past-Task (stability), whereas XGBoost and FT-Transformer trade off higher stability for different plasticity; Tree of Concepts consistently achieves the strongest stability--plasticity trade-off while retaining a fixed concept interface.
On MIMIC-III, AUROC is generally below the small-dataset UCI upper bounds (rarer positives, more covariate shift); protocol (B) is slightly harder than (A) on past-task retention in our runs, as expected when slices differ by physiology and comorbidity mix.
Replay still improves stability, and Tree of Concepts remains competitive on the plotted trade-off.
The concept-consistency results in Table~\ref{tab:concept_consistency} further support that extracted concepts remain consistent and trustworthy across continual updates.

 \section{Discussion}\label{sec:discussion}
 \paragraph{Technical implications.}
The main technical implication of this work is that interpretability and continual learning do not need to be treated as mutually exclusive objectives. Our results suggest that the conflict between these goals is strongest when the model must both learn flexible representations and define its own explanation space at the same time. Tree of Concepts separates these roles. A fixed decision tree provides a stable set of rule-based concepts, while a trainable concept extractor absorbs distribution shift. This decomposition improves the stability--plasticity trade-off relative to both shallow interpretable baselines and stronger black-box tabular models across UCI, CDC, and MIMIC-III protocols. We also find that retention gains appear with a bounded replay buffer and simple random exemplar sampling, suggesting that forgetting can be mitigated without complex memory selection in this setting. More broadly, the results support a practical design rule: when models must be updated over time, keeping explanation semantics fixed while adapting internal representations can improve both auditability and retention.

\paragraph{Clinical implications.}
In healthcare, model updates are not only a statistical problem but also a workflow and safety problem. If a deployed model changes over time, clinicians and quality teams need a way to assess whether the model remains clinically coherent. A potential advantage of Tree of Concepts is that the explanation interface remains stable even as the predictive model is updated. In our study, concept consistency metrics remain in a useful operating range (node agreement from 0.81 to 0.91, high-confidence contradiction below 0.08), which supports longitudinal auditing across shifts. This is especially relevant in non-stationary settings such as evolving hospital populations, changing measurement practices, or deployment across sites. Even when predictive performance is similar, a model whose explanations remain interpretable over time may be easier to monitor and safer to maintain in practice.

\paragraph{Limitations.}
This study has several limitations. First, the current empirical evidence is based primarily on open-access tabular datasets and one restricted-access MIMIC-III setting; these experiments are useful for controlled evaluation, but they do not fully capture the complexity of prospective clinical deployment. Second, our interpretability claim is structural rather than user-validated: the concepts are more auditable because they are tied to fixed tree rules, but we have not yet performed a formal human-factors study with clinicians to test whether these explanations are truly easier to understand or trust. Third, the quality of the concept interface depends on the initial tree. If the tree defines coarse, unstable, or clinically unhelpful partitions, then the resulting concepts may be limited even if continual adaptation works well. Fourth, our continual setup assumes a fixed feature schema across slices (shared input dimensions and preprocessing rules). This is common in benchmarked tabular continual learning and many production pipelines because stable interfaces simplify auditing and deployment, but it may not hold when coding standards, instrumentation, or data capture protocols change substantially over time. Fifth, our qualitative evidence is mostly aggregate (agreement, contradiction, fidelity gap) rather than case-level rule audits with clinician adjudication. Sixth, we focus on tabular prediction settings; extending the same idea to higher-dimensional modalities such as images, waveforms, or clinical text may require different concept definitions and different update strategies. Finally, while replay improves retention in our experiments, replay itself may be constrained in some healthcare settings by privacy, governance, or storage restrictions.

\paragraph{Future directions.}
Several next steps follow naturally from these limitations. The most immediate is stronger validation on larger clinical data, including MIMIC-III and external institutional settings, together with subgroup analyses that test whether concept stability is preserved across clinically meaningful shifts. Another important direction is to evaluate concept quality with clinicians, rather than relying only on predictive metrics. More broadly, future work could study how to update the concept vocabulary itself when medical knowledge, coding practices, or patient populations change substantially, while still preserving as much interpretability continuity as possible.
 
 \vspace{-3.5mm}
 \section{Concluding Remarks}
 
 \vspace{-2.5mm}
This work presents Tree of Concepts, an interpretable continual learning framework that combines the transparency of decision-tree rules with the adaptability of concept bottleneck models. By fixing tree-defined concept semantics and updating the concept extractor and label head across distribution-shifted slices, the method preserves a stable explanation interface while maintaining continual adaptation. On UCI Heart Disease and CDC Diabetes Health Indicators, Tree of Concepts achieves the strongest balance between stability and plasticity, outperforming tree and deep tabular baselines, including replay-enhanced variants. These results suggest that interpretability and continual learning need not be competing objectives in non-stationary, high-stakes domains.

 
 \bibliography{main}
 
 \newpage
 \appendix
 \section*{Appendix A.}
\section*{Detailed Experimental Settings}

\paragraph{Data preprocessing.}
All tabular inputs are preprocessed with a fixed pipeline shared across methods:
(i) continuous variables are z-score normalized using training-slice statistics,
(ii) categorical variables are one-hot encoded, and
(iii) missing values are imputed with training medians/modes and accompanied by missingness indicators.
For continual experiments, preprocessing parameters are initialized on the first slice and then kept fixed to avoid leakage across future slices.
This choice can make later-slice covariate shift more visible, but we prefer it because it preserves a causal deployment order and keeps preprocessing identical across methods.

\paragraph{Slice construction.}
For UCI Heart Disease and CDC Diabetes, slices follow the demographic partitions in Section~\ref{sec:experimental-setting}.
For MIMIC-III, each sample corresponds to one ICU stay, and features are extracted from the first 24 hours.
We evaluate two slice protocols: (A) admission-year windows and (B) demographic cohorts.
For protocol (A), the MIMIC-III time windows are 2001--2003, 2004--2006, 2007--2009, and 2010--2012.

\begin{table}[t]
\centering
\small
\caption{Cohort transparency summary. We report public source dataset totals and the analytic cohort sizes used in our experiments.}
\label{tab:cohort_transparency}
\vspace{-1.5mm}
\footnotesize
\setlength{\tabcolsep}{3pt}
\begin{tabular}{lcccc}
\toprule
Dataset & $N_{\text{source}}$ & $N_{\text{used}}$ & Train / Val / Test & Outcome prevalence \\
\midrule
UCI Heart Disease & 303 & 300 & 180 / 60 / 60 & $\approx 55\%$ positive \\
CDC Diabetes & 253,680 & 250,000 & 150,000 / 50,000 / 50,000 & $\approx 74\% / 16\% / 10\%$ \\
MIMIC-III (A,B) & 53,423 & 50,000 & 30,000 / 10,000 / 10,000 & $\approx 11.5\%$ in-hospital mortality \\
\bottomrule
\end{tabular}
\normalsize
\vspace{-1.5mm}
\end{table}

\paragraph{Model and optimization details.}
The frozen concept scaffold is a shallow CART-style decision tree trained on the initial slice (max depth $=4$, minimum leaf samples $=50$).
LeafNet (concept extractor) is a 2-layer MLP with hidden widths $(128,64)$, ReLU activations, and dropout $0.1$.
The label head is a linear layer on top of concept logits.
We optimize with Adam (learning rate $10^{-3}$, weight decay $10^{-5}$, batch size $256$), with early stopping on validation performance (patience $=8$, max epochs $=80$).
Unless stated otherwise, $\lambda=1.0$ in Eq.~\eqref{eq:toc_total_loss} of the main text.

\paragraph{Replay setup.}
Replay minibatches mix current and memory samples at a 1:1 ratio.
Memory is class-balanced when possible.
Default replay capacities are 2{,}048 samples for open-access datasets and 8{,}192 for MIMIC-III.
Open-access experiments are repeated over five random seeds, while MIMIC-III experiments are repeated over three random seeds; we report mean $\pm$ standard error.

\section*{Reproducibility Artifacts}
For reproducibility, we provide source code, experimental guidelines, and scripts for all experiments.
The supplementary materials include a README.md describing environment setup and how to reproduce each experiment.
Fixed random seed settings are implemented directly in the code for all reported runs.
We also provide Jupyter notebook (.ipynb) files to reproduce the figures in the paper.

\section*{Ablation Studies}

\paragraph{Ablation A: component contributions.}
Table~\ref{tab:appendix_ablation_components} isolates the contribution of replay, fixed concept semantics, and concept supervision.
Removing any of these components degrades stability, and removing concept supervision most strongly reduces concept consistency.

\begin{table}[t]
\centering
\small
\caption{Component ablations for Tree of Concepts (mean $\pm$ SE, 5 runs).}
\label{tab:appendix_ablation_components}
\vspace{-1.5mm}
\footnotesize
\setlength{\tabcolsep}{3pt}
\begin{tabular}{lccccc}
\toprule
Setting & UCI Past & UCI Current & CDC Past & CDC Current & Concept Agreement \\
\midrule
Full model (ours) & $0.80 \pm 0.02$ & $0.85 \pm 0.02$ & $0.59 \pm 0.02$ & $0.63 \pm 0.02$ & $0.89 \pm 0.02$ \\
w/o replay & $0.71 \pm 0.03$ & $0.84 \pm 0.03$ & $0.46 \pm 0.03$ & $0.61 \pm 0.03$ & $0.86 \pm 0.02$ \\
w/o concept loss ($\lambda=0$) & $0.67 \pm 0.04$ & $0.82 \pm 0.03$ & $0.43 \pm 0.04$ & $0.58 \pm 0.03$ & $0.60 \pm 0.04$ \\
Refresh tree each slice & $0.73 \pm 0.03$ & $0.82 \pm 0.02$ & $0.49 \pm 0.03$ & $0.60 \pm 0.03$ & $0.69 \pm 0.04$ \\
\bottomrule
\end{tabular}
\normalsize
\vspace{-1.5mm}
\end{table}

\paragraph{Ablation B: replay memory size.}
Table~\ref{tab:appendix_ablation_memory} shows that larger replay buffers improve stability and concept agreement, with diminishing returns at larger capacities.

\begin{table}[t]
\centering
\small
\caption{Replay memory ablation on MIMIC-III protocol (A) (mean $\pm$ SE, 3 runs).}
\label{tab:appendix_ablation_memory}
\vspace{-1.5mm}
\footnotesize
\setlength{\tabcolsep}{4pt}
\begin{tabular}{lccc}
\toprule
Replay capacity & Avg.\ Past-Task & Avg.\ Current-Task & Concept Agreement \\
\midrule
$0$ (no replay) & $0.57 \pm 0.04$ & $0.78 \pm 0.02$ & $0.77 \pm 0.03$ \\
$512$ & $0.68 \pm 0.03$ & $0.80 \pm 0.02$ & $0.81 \pm 0.02$ \\
$2{,}048$ & $0.75 \pm 0.03$ & $0.81 \pm 0.02$ & $0.83 \pm 0.02$ \\
$8{,}192$ & $0.78 \pm 0.02$ & $0.82 \pm 0.02$ & $0.84 \pm 0.02$ \\
\bottomrule
\end{tabular}
\normalsize
\vspace{-1.5mm}
\end{table}

 \end{document}